\documentclass[10pt,twocolumn,letterpaper]{article}
\pdfoutput=1

\usepackage[pagenumbers]{iccv} 
\usepackage{times}
\usepackage{epsfig}
\usepackage{graphicx}
\usepackage{amsmath}
\usepackage{amssymb}
\usepackage{subcaption}
\usepackage{float}

\usepackage{MnSymbol,bbding,pifont}
\usepackage{tcolorbox}
\usepackage{siunitx}
\usepackage{hhline}
\usepackage[normalem]{ulem} 
\usepackage{xcolor}
\usepackage{booktabs}
\usepackage{makecell}
\usepackage{tabularx} 
\usepackage{colortbl}
\usepackage{comment}
\usepackage{enumitem}
\usepackage{CJKutf8}
\usepackage{graphicx}
\usepackage{algorithm}
\usepackage{algorithmic}
\usepackage{multirow}
\usepackage{appendix} 
\usepackage[accsupp]{axessibility}  

\newcommand{\VarSty}[1]
{\textbf{\ttfamily\color{blue!80!black}#1}\unskip}

\newcommand{\var}{\texttt}

\definecolor{iccvblue}{rgb}{0.21,0.49,0.74}
\usepackage[pagebackref,breaklinks,colorlinks,allcolors=iccvblue]{hyperref}

\def\paperID{xxxx} 
\def\confName{ICCV}
\def\confYear{2025}

\title{When Large Vision-Language Model Meets Large Remote Sensing Imagery: Coarse-to-Fine Text-Guided Token Pruning}

\author{\textbf{Junwei Luo}$^{1}$\thanks{indicates interns at Ant Group.  $^{\dagger}$Corresponding author.}\ , \ 
\textbf{Yingying Zhang}$^{2}$, \ 
\textbf{Xue Yang}$^{3}$, \ 
\textbf{Kang Wu}$^{1}$\footnotemark[1]\ , \ 
\textbf{Qi Zhu}$^{2}$, \ \\
\textbf{Lei Liang}$^{2}$, \ 
\textbf{Jingdong Chen}$^{2}$,\ 
\textbf{Yansheng Li}$^{1\dagger}$\\
$^{1}$Wuhan University \quad 
$^{2}$Ant Group \quad  $^{3}$SAIS, Shanghai Jiao Tong University\\
{\tt\small \{luojunwei,kangwu,yansheng.li\}@whu.edu.cn} \quad 
{\tt\small yangxue-2019-sjtu@sjtu.edu.cn} 
  \\
{\tt\small  \url{https://github.com/VisionXLab/LRS-VQA}}
}

\begin{document}
\maketitle

\begin{abstract}
  Efficient vision-language understanding of large Remote Sensing Images (RSIs) is meaningful but challenging. Current Large Vision-Language Models (LVLMs) typically employ limited pre-defined grids to process images, leading to information loss when handling gigapixel RSIs. Conversely, using unlimited grids significantly increases computational costs. To preserve image details while reducing computational complexity, we propose a text-guided token pruning method with Dynamic Image Pyramid (DIP) integration. Our method introduces: (i) a Region Focus Module (RFM) that leverages text-aware region localization capability to identify critical vision tokens, and (ii) a coarse-to-fine image tile selection and vision token pruning strategy based on DIP, which is guided by RFM outputs and avoids directly processing the entire large imagery. Additionally, existing benchmarks for evaluating LVLMs' perception ability on large RSI suffer from limited question diversity and constrained image sizes. We construct a new benchmark named LRS-VQA, which contains 7,333 QA pairs across 8 categories, with image length up to 27,328 pixels. Our method outperforms existing high-resolution strategies on four datasets using the same data. Moreover, compared to existing token reduction methods, our approach demonstrates higher efficiency under high-resolution settings.

\end{abstract}

\vspace{-5pt}

\section{Introduction}

Benefiting from the rapid advancement of Large Language Models (LLMs)~\cite{achiam2023gpt,vicuna2023,dubey2024llama,qwen2.5,deepseekai2025deepseekr1}, Large Visual-Language Models (LVLMs) have demonstrated strong capabilities in perceiving and understanding visual information in text-based multimodal interactions ~\cite{liu2023improved,li2024llava,wang2024qwen2vl,internvl25}. Currently, LVLMs have been extensively studied and applied across various fields like remote sensing (RS) intelligent interpretation~\cite{muhtar2024lhrs,zhang2024earthgpt,zhu2025skysenseo,xue2024reo,li2025lmmrotate,danish2024geobenchvlm}.

\begin{figure}[!t]
    \centering
    \includegraphics[width=\columnwidth]{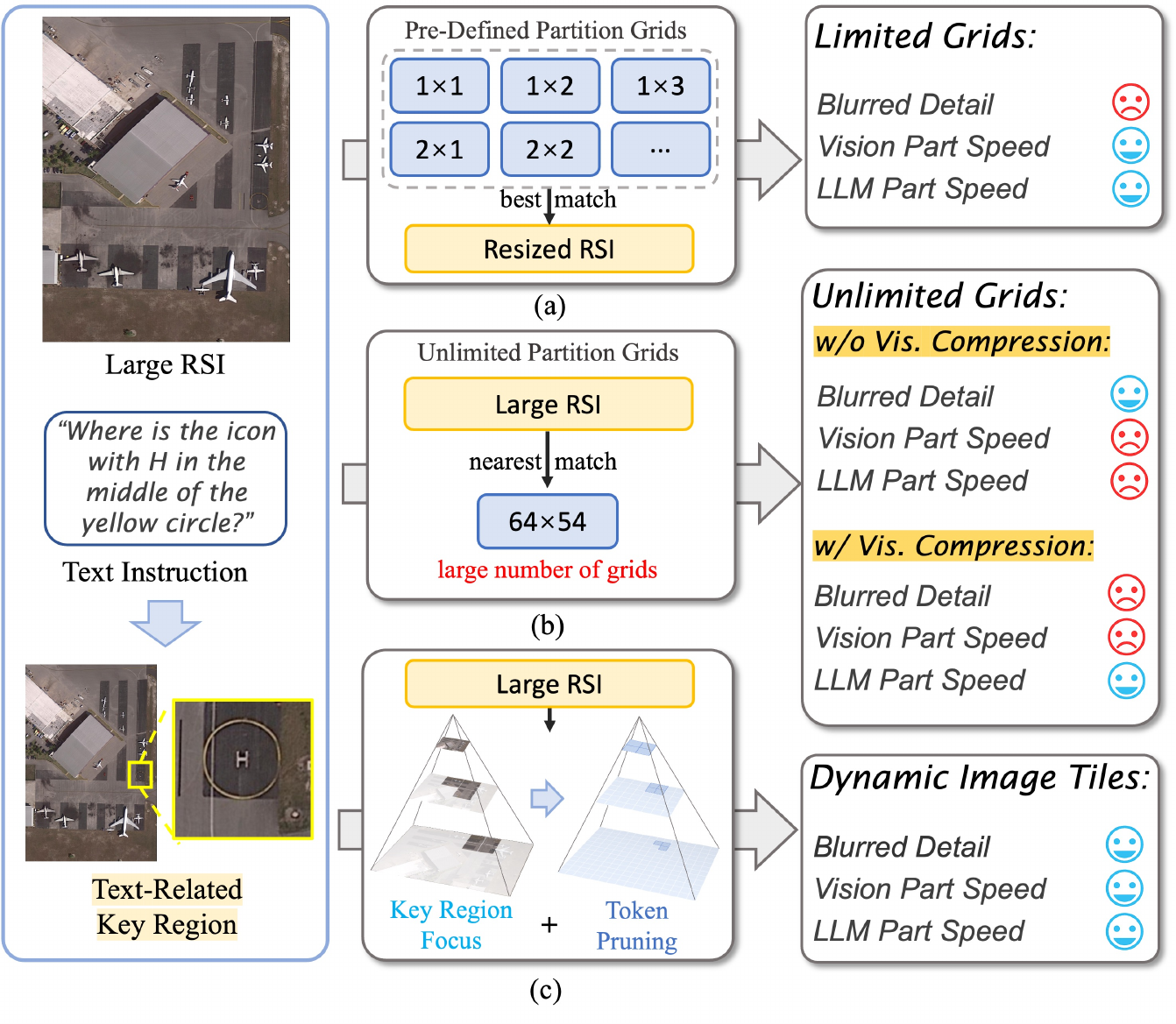}
    \vspace{-13pt}
    \caption{High-resolution strategy comparison for modular LVLMs. (a) and (b) show that existing grid-based cropping methods face challenges when processing large RSIs. (c) The proposed dynamic pyramid-based token pruning strategy can dynamically select image tiles of key regions related to the input text, balancing image detail and computational cost.}
    \vspace{-12pt}
    \label{fig:introduction}
\end{figure}

Advances in satellite imaging technology allow for the acquisition of large remote sensing images (RSIs) that cover extensive ground areas and contain detailed land use information~\cite{guo2024skysense,li2025meet}. Therefore, language-based intelligent analysis of large RSIs is valuable for applications such as complex scene understanding~\cite{li2024scene,luo2024skysensegpt}, urban planning~\cite{wang2024earthvqa,li2024learning}, infrastructure development~\cite{,xiao2024foundation}, and monitoring~\cite{zhou2024vlgfm}.

Some RS LVLMs handle high-resolution input through position embedding interpolation~\cite{kuckreja2023geochat,zhang2024earthmarker}. Additionally, various dynamic high-resolution strategies designed for general LVLMs~\cite{ye2023ureader,liu2023improved,li2024tokenpacker} can be adapted to RS LVLMs, such as EarthDial~\cite{soni2024earthdial} (based on InternVL1.5~\cite{chen2024internvl15}) and GeoPixel\cite{shabbir2025geopixel} (using InternLM-XComposer2.5~\cite{internlmxcomposer2_5}). However, these approaches for high-resolution image processing are limited when handling large RSIs exceeding 10,000$\times$10,000 pixels. As depicted in Fig.~\ref{fig:introduction}, these methods typically employ limited pre-defined grid partitions, leading to detail loss when large RSIs are excessively downsampled. Conversely, unlimited grids suffer from prohibitive time and memory costs (e.g., over 5 million vision tokens for a large RSI in LLaVA-1.5~\cite{liu2023improved}). This necessitates balancing resolution and computational efficiency when processing large RSI.

To overcome the above limitations of existing strategies in handling large RSIs, we propose a text-guided token pruning strategy that consists of two key components: a Region Focus Module (RFM) and a Dynamic Image Pyramid (DIP). The RFM identifies text-relevant key vision tokens by leveraging the capability distilled from the LLM component. Based on the RFM's output, we select critical image tiles and conduct token pruning in a coarse-to-fine manner within the DIP. This enables the LVLM to focus and zoom in key areas, avoiding to process all visual tokens, thereby facilitating flexible and efficient inference. 

Moreover, benchmarks for LVLMs in understanding large RSIs remain insufficiently developed. Although MME-Realworld~\cite{zhang2024mme} includes large RSIs with high-quality manually annotated questions, its RS subset suffers from limited question diversity and image sizes. To more comprehensively characterize the challenges of large RSI perception, we construct a new benchmark called LRS-VQA (\textbf{L}arge \textbf{R}emote \textbf{S}ensing image Visual Question Answering). LRS-VQA features larger size images with lengths up to 27,328 pixels and 8 distinct question types. Our key contributions are as follows:

\begin{itemize}[itemsep=0pt, parsep=0pt]
\item We introduce a Region Focus Module (RFM) that distills the ability from LLM part of LVLM to efficiently identify text-related key vision tokens.

\item Based on RFM and DIP, we propose an efficient coarse-to-fine inference strategy for large RSIs, which achieves a balance between resolution and efficiency through key tile selection and vision token pruning.

\item   We construct a new benchmark named LRS-VQA, featuring larger image sizes and more diverse question types than existing benchmarks, to reflect the challenges of large RSIs perception.

\item The proposed method is architecture-agnostic and shows performance improvements and efficiency gains.
\end{itemize}

\begin{figure*}[tbp]
    \centering
    \includegraphics[width=\textwidth]{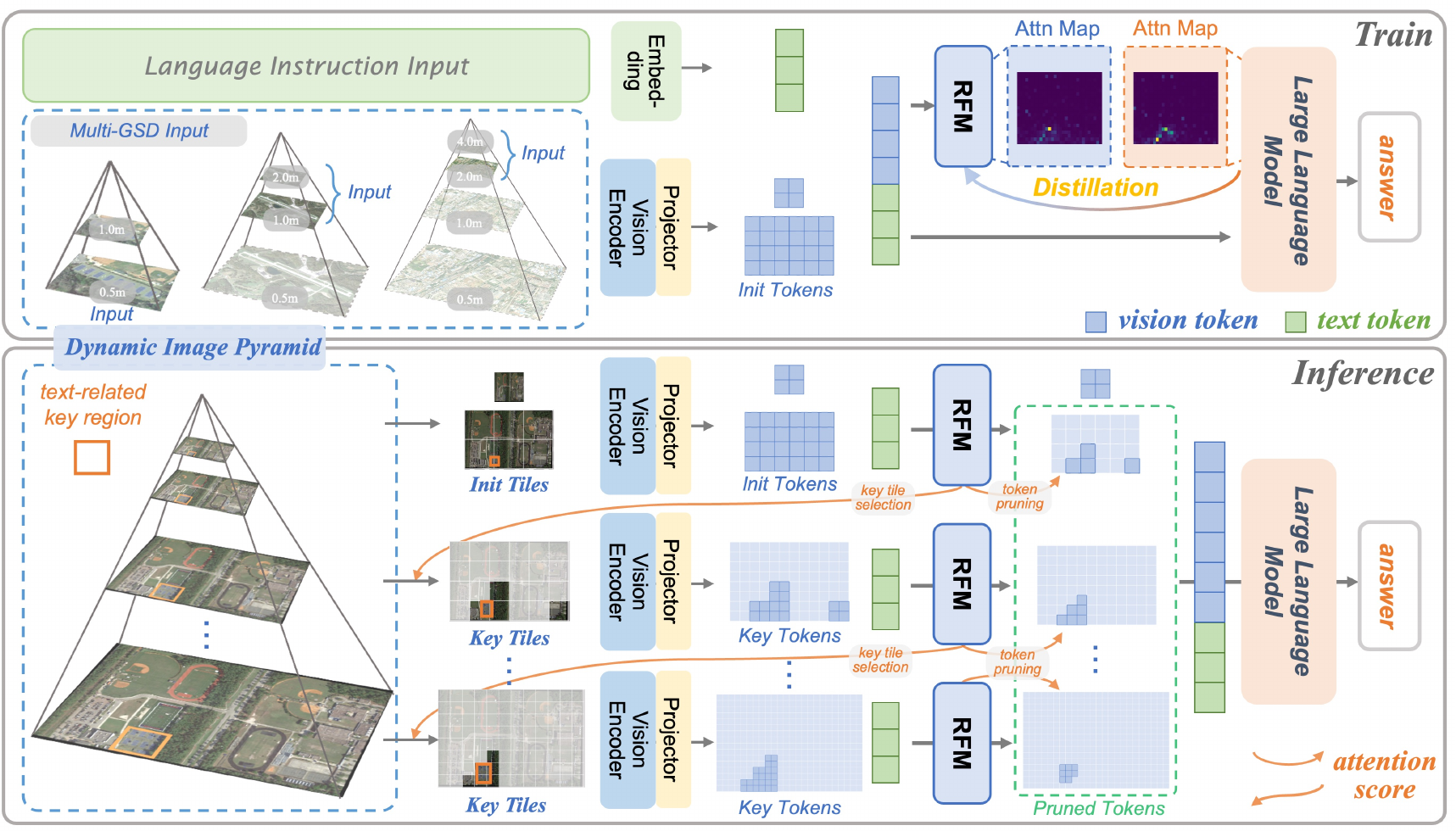}
    \vspace{-15pt}
    \caption{The pipeline of the proposed method. The entire process iterates in a coarse-to-fine manner, dynamically retrieving high-resolution features from the next DIP level (leftward orange arrow) or performing token pruning at the current level (rightward orange arrow) based on the output of the RFM module at each iteration. During training, the RFM distillation text-related attention from the LLM; during inference, RFM generates the attention scores for the input vision tokens. GSD means ground sample distance.}
    \label{fig:pipeline}
    \vspace{-8pt}
\end{figure*}
\section{Related Work}

\subsection{LVLMs for High-Resolution Comprehension}

Numerous LVLMs designed for high-resolution image or video perception have recently emerged. They can be primarily categorized into the following paradigms: 

\textbf{Methods that carefully design image cropping and padding strategy.} Methods like ~\cite{ye2023ureader,liu2023improved,li2024llava,li2024monkey,liu2024textmonkey,chen2024internvl15,guo2025llavauhd,zhang2024slime} traverse pre-defined partition grids and obtain the most suitable grid, to divide the high-resolution image into image tiles. These methods encounter limitations when processing large RSIs as in Fig.~\ref{fig:introduction}.

\textbf{Methods based on multiple visual branches.} These approach~\cite{li2024mgm,tong2025cambrian,luo2024feast} employ additional high-resolution visual encoders like SAM~\cite{kirillov2023segment} encoder or ConvNext~\cite{liu2022convnet} to process images with higher input resolution. 

\textbf{Visual chain-of-thoughts methods.} These approaches utilize a search-and-focus strategy~\cite{wu2024vstar,shao2025visual,hrbench,shen2024zoomeye} or external RAG pipeline~\cite{zhang2024enhancing} to identify text-related key image regions. However, such pipelines require multiple LVLM inferences, making them more complex and cumbersome.

\subsection{Vision Token Reduction in LVLMs}

Visual token pruning for transformers has been a classic research topic, it aims to accelerate computation by dynamically reducing less important tokens.  It has been extensively studied in both natural language processing~\cite{ye2021tr,kim2022learned} and computer vision domains~\cite{rao2021dynamicvit,meng2022adavit,kong2022spvit,wei2023joint,dong2023heatvit}.

Recently, a variety of token pruning methods have been proposed for LVLMs, which can be categorized into three main types: token pruning conducted in the vision encoder ~\cite{yao2024deco,shang2024llavaprumerge,li2024tokenpacker,arif2024hired,wang2024longllava,yang2024visionzip}, token pruning performed in the LLM component~\cite{chen2024fastv,xing2024pyramiddrop,ye2024fit,han2024rethinking,zhang2024sparsevlm,huang2024hires}, and collaborative approaches that integrate both~\cite{li2023blip,llavamini,bai2023qwen,li2024flexattention,li2024llamavid,yan2024tgllava,ye2024voco}. Moreover, adaptive token segmentation methods~\cite{lew2024superpixel,aasan2024spitting,chen2024subobject} also offer a promising approach for token reduction.

Although vision-part pruning methods efficiently reduce vision tokens, when processing large images, it's highly time-consuming to traverse pre-defined grids and handle numerous image tiles with the vision encoder. Moreover, the absence of language guidance makes it hard to identify foreground regions in complex RSIs. The LLM-part pruning methods require feeding all vision tokens into the LLM. Therefore, long visual sequences from large images may exceed LLM length limits and incur high computational costs for attention score ranking.

Among the collaborative approaches, the methods most closely related to ours are LLaVA-Mini~\cite{llavamini} and FlexAttention~\cite{li2024flexattention}. The former employs modality pre-fusion to compress visual information; however, its LLM-like pre-fusion module still struggles with handling extremely long vision sequences. The latter integrates text-guided high-resolution features within the LLM component. Nevertheless, the allowable image size for its high-resolution view remains limited, and it can only index high-resolution features once.

\section{Preliminaries}

In this section, we present the foundational concepts of text-image attention within the LLM component of LVLMs. Mainstream modular LVLMs include three components~\cite{luo2024mono}: a pre-trained vision encoder, a projector, and a pre-trained decoder-only LLM. Conventional high-resolution strategies~\cite{chen2024internvl15,guo2025llavauhd} process the original image $I_{img}$ into a thumbnail view $I_\text{thumb}$ and a set of image tiles $I_\text{tiles}$. Then the vision encoder and the projector transforms $I_\text{thumb}$ and $I_\text{tiles}$ into vision token embeddings $T_{vis}=[T^{lr}_{vis}, T^{hr}_{vis}]$, which contain low-resolution tokens $T^{lr}_{vis}$ from the $I_\text{thumb}$ and higher-resolution $T^{hr}_{vis}$ from the $I_\text{tiles}$. $T_{vis}$ are concatenated with text token embeddings $T_{txt}$ from the tokenized text instruction, forming a multimodal token sequence $T=[T_{vis}, T_{txt}]$. The LVLM processes $T$ via transformer-based decoder layers with causal self-attention, where the attention score $A$ can be represented as:
\begin{equation}
A = \text{Attention}(Q, K) = \text{Softmax}\left(\frac{Q K^\top}{\sqrt{d}}\right),
\label{eq:self_attention}
\end{equation}
where $A\in \mathbb{R}^{n \times n}$ is the self-attention map, $Q, K\in \mathbb{R}^{n \times d}$ are the query and key matrices derived from the input embeddings through linear transformations, $n$ is the input sequence length, and $d$ is the feature dimension. As $A$ determines the weight each token assigns to the former tokens in the sequence, by analyzing the attention from text tokens to vision tokens, we can identify text-related important vision tokens. Some studies~\cite{kaduri2024s,ye2024fit} observe that in the deep layers of LLM part in LVLMs, attention is predominantly allocated to text-related vision tokens compared to other vision tokens. This insight inspired us to design a coarse-to-fine mechanism to focus on key image regions.

\begin{figure*}[!t]
    \centering
    \includegraphics[width=\textwidth]{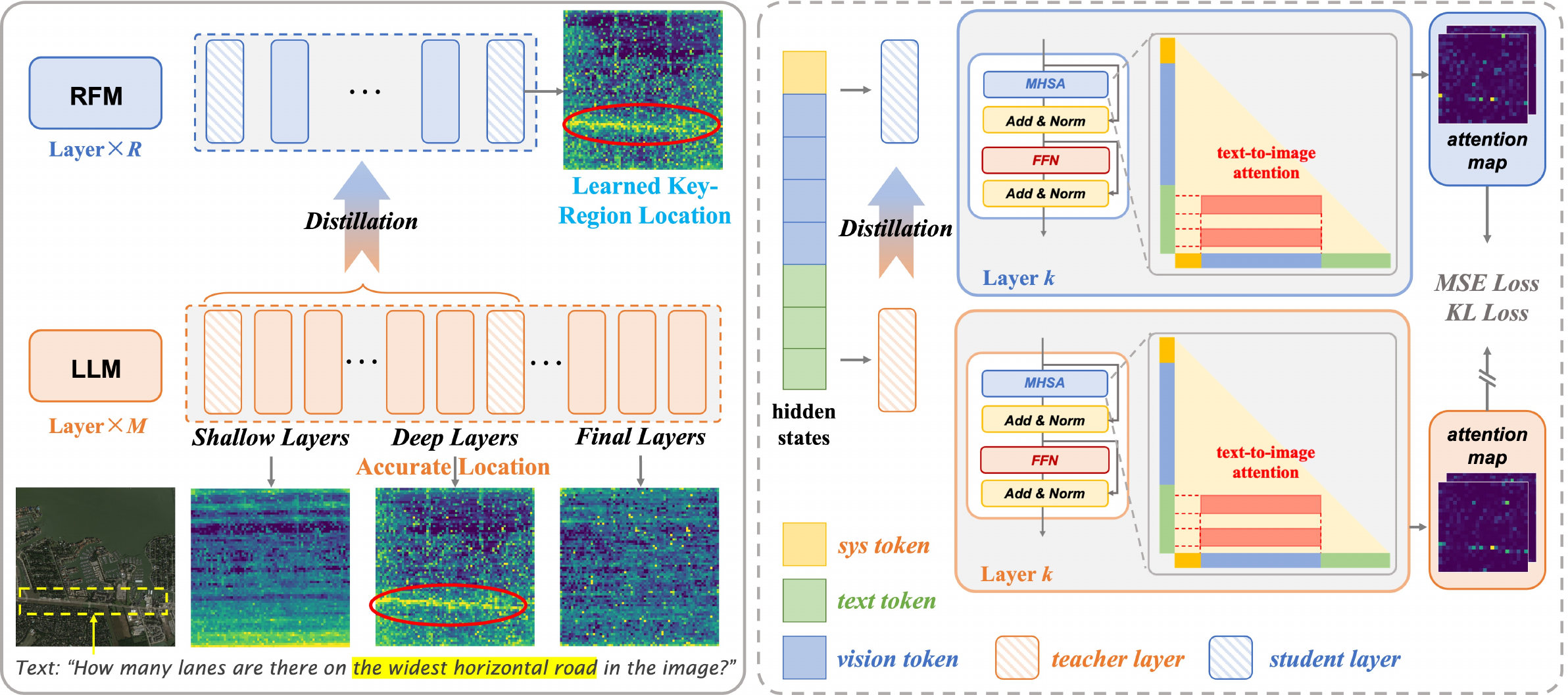}
    \vspace{-14pt}
    \caption{The proposed RFM and attention distillation strategy. The left part indicates our core idea: distill accurate text-related key region localization ability from the LLM part of the LVLM. The right part shows the distillation details. We only select specific layer pairs for distillation to avoid hidden state discontinuities. ``sys token" represents the tokens from the system prompt.}
    \vspace{-12pt}
    \label{fig:distill}
\end{figure*}

\vspace{-5pt}

\section{Method and Benchmark}

\subsection{Method}

To achieve efficient understanding of ultra-high-resolution images, our core idea is \textbf{coarse-to-fine focusing}: we first gain a low-resolution overview, then iteratively zoom into text-related regions for details, as shown in Fig.~\ref {fig:pipeline}. Implementing this necessitates a multi-resolution representation. Therefore, we first introduce the Dynamic Image Pyramid (DIP) in Sec.~\ref{sec:DIP}. Then we need to identify where to focus next. To solve this, we propose the Region Focus Module (RFM) to provide attention distribution for input vision tokens (Sec.~\ref{sec:RFM}). Finally, these components are integrated into an iterative refinement process (Sec.~\ref{sec:pruning}).

\subsubsection{The Construction of Dynamic Image Pyramid}
\label{sec:DIP}

As established, our coarse-to-fine strategy requires a multi-resolution representation of the image $I_{img}$. We realize this through the DIP, a structure designed to provide image tiles at various ground sample distances (GSDs).

First, we iteratively downsample $I_{img}$ by a factor of 2, generating a series of images $\{I^1_{init}, I^2_{init}, \dots, I^{P}_{init}\}$ until the shorter side reaches a pre-defined minimum length. Let a basic image tile size be $B \times B$ (e.g., $336 \times 336$ for CLIP-L14), for the $p$-th scaled image $I^{p}_{init}$, we calculate the number of image tiles along the height and width as follows:
\vspace{-8pt}
\begin{align}
N_h^p = \lceil H_p / B \rceil, \quad N_w^p = \lceil W_p / B \rceil,
\end{align}
where $H_p, W_p$ are the height and width of $I^{p}_{init}$, respectively. For proper tiling, we compute a scale factor $r_p$ as:
\vspace{-8pt}
\begin{align}
    r_p &= \min\left(\frac{N_h^p \times B}{H_p}, \frac{N_w^p \times B}{W_p}\right).
\end{align}

The image $I^{p}_{init}$ is then resized to $I^{p}$ using $r_p$, followed by padding to preserve the original aspect ratio. After resizing, $I^{p}$ is partitioned into non-overlap tiles $I^{p}_\text{tiles}$. Tiles from all scaled images are combined in reverse order, and integrated with the thumbnail view $I_\text{thumb}$ to form the final pyramid: $ I_{DIP}=\{I_\text{thumb}, I^1_\text{tiles}, I^2_\text{tiles}, \dots, I^{\text{P}}_\text{tiles}\}$, which consists of $P+1$ layers with progressively increasing resolutions. During training, we select the thumbnail $I_{\text{thumb}}$ and $I^1_{\text{tiles}}$ as visual inputs for computational efficiency. During inference, all pyramid layers become accessible to enable multi-scale inference, as in the left part of Fig.~\ref{fig:pipeline}.

\subsubsection{Attention Distillation for Key Region Focus}
\label{sec:RFM}
With the DIP established, the central challenge becomes generating a reliable guidance signal to determine where to focus next. While a sliding-window similarity map from VLMs like CLIP is a straightforward baseline, we posit that the attention maps from the LVLM's LLM part provide a superior localization signal. This is inspired by prior work~\cite{kaduri2024s,xing2024pyramiddrop,ye2024fit} where LLM attention tends to converge on key visual tokens. To support this, we quantitatively validate that LLM attention significantly outperforms the CLIP/RemoteCLIP baselines by offering more accurate, text-related localization signals in Tab.~\ref{tab:location_acc}.

To harness this high-quality guidance efficiently, we introduce the Region Focus Module (RFM), a lightweight module trained via attention distillation to mimic the text-guided localization capability of the LLM teacher, as shown in Fig.~\ref{fig:distill}. After distillation, the trained RFM can guide the selection of key image tiles and prune irrelevant vision tokens before the main LLM computation.

Specifically, assuming the LLM has $M$ layers $l_1,l_2,\dots, l_M$, the proposed RFM adopts the same architecture with $R$ layers $r_1,r_2,\dots, r_R$, where $R$$<$$M$. Each RFM layer $r_i$ is initialized from a selected LLM layer $l_{m_i}$, where $\{m_1, m_2, \dots, m_R\}$ forms a sparse subset of LLM layer indices. For layer-wise distillation, we select $K$ pairs ($K$$<$$R$) from RFM and LLM as student-teacher pairs, represented as $\{(r_k, l_{m_k}) \mid k = 1, 2, \dots, K\}$. This strategy avoids rigidly aligning each $r_i$ with $l_{m_i}$, which would neglect the non-teacher layers in the LLM, resulting in inconsistency between adjacent RFM layers.

As depicted in the right part of Fig.~\ref{fig:distill}, the concatenated multimodal tokens $T=[T_{vis}, T_{txt}]$ are fed into both the RFM and the LLM during training. For the $k$-th student-teacher layer pair $(r_k, l_{m_k})$, we extract the attention scores between the last text token and all vision tokens from the multi-head self attention (MHSA). For the $h$-th head, the self-attention is represented as $A^{k,h}_{\text{stu}}, A^{k,h}_{\text{tea}}\in \mathbb{R}^{j \times n_{v}}$, where $j$ is the number of dialogue turns for multi-turn instruction and $n_{v}$ is the length of vision tokens. We apply the Kullback-Leibler (KL) divergence loss for distillation:
\vspace{-8pt}
 \begin{equation}
\mathcal{L}^{h}_{\text{kl}} =\sum_{k=1}^{K} \left[ D_{\text{KL}}\left(A^{k,h}_{\text{tea}} \,\|\, A^{k,h}_{\text{stu}} \right) + \lambda_{\text{hr}} D_{\text{KL}}\left( A^{k,h,\text{hr}}_{\text{tea}} \,\|\,  A^{k,h,\text{hr}}_{\text{stu}} \right)
\right],
\label{eq:kl_loss}
\end{equation}
where $A^{k,h,\text{hr}}_{\text{tea}}, A^{k,h,\text{hr}}_{\text{stu}}$ denote attention corresponding to $T^{hr}_{vis}$, and $\lambda_{\text{hr}}$ is the weighting factor. Additionally, we enforce stronger alignment constraints to the attention of $T^{hr}_{vis}$ by using an additional Mean Squared Error (MSE) loss:
\vspace{-7pt}
\begin{equation}
\mathcal{L}^{h}_{\text{mse}} = \sum_{k=1}^{K} \text{MSE}\left( A^{k,h,\text{hr}}_{\text{stu}},\  A^{k,h,\text{hr}}_{\text{tea}} \right).
\label{eq:mse_loss}
\end{equation}
\vspace{-0.6pt}

The total distillation loss is computed as:
\vspace{-7pt}
\begin{equation}
\mathcal{L}_{\text{distill}} = \frac{1}{KH} \sum_{h=1}^{H} \left( \lambda_{\text{mse}}  \mathcal{L}^{h}_{\text{mse}} + \lambda_{\text{kl}}   \mathcal{L}^{h}_{\text{kl}} \right),
\label{eq:total_loss}
\end{equation}
where $K$ is the number of selected student-teacher layer pairs, $H$ is the number of attention heads, and $\lambda_{\text{mse}}, \lambda_{\text{kl}}$ are hyperparameters. For the flash-attention~\cite{dao2022flashattention}, we employ a specialized value matrix to extract attention maps like~\cite{zhang2024sparsevlm} to ensure compatibility.

\subsubsection{Text-Guided Token Pruning with Pyramid}
\label{sec:pruning}

Having introduced the DIP and the RFM, this section details how they are integrated into a cohesive, iterative process. This process achieves our coarse-to-fine strategy by selectively retrieving high-resolution tiles and pruning irrelevant tokens, as illustrated in Fig.~\ref{fig:pipeline}.

The specific inference procedure, detailed in Alg.~\ref{alg:pyramid_construction}, is as follows. For simplicity, the tokens from the system prompt are omitted. We initialize the vision input using $I_\text{thumb}$ and $I^p_\text{tiles}$ to generate the initial $T_{vis} = [T^{lr}_{vis}, T^{p, hr}_{vis}]$, where $p=1$, then the concatenated tokens $T$ are fed into the RFM. From the last layer of the RFM, we compute the average attention scores from all heads in the MHSA, and extract the attention scores $\smash{A^{K, hr}_{\text{stu}}}$ of $\smash{T^{p, hr}_{\text{vis}}}$. By selecting the top-$\lambda$ proportion from $\smash{A^{K, hr}_{\text{stu}}}$, we identify the indices of key tokens and map their coordinates to image tile-level coordinates, to select key image tiles $I^{p+1}_\text{key}$ from $I^{p+1}_\text{tiles}$. Based on the number of $I^{p+1}_\text{key}$, we determine whether to prune the $\smash{T^{p, hr}_{\text{vis}}}$ directly or replace it with vision tokens $\smash{T^{p+1, hr}_{\text{vis}}}$ from higher-resolution tiles. If $\smash{T^{p+1, hr}_{\text{vis}}}$ is obtained, the tile selection process is repeated recursively until the last layer of DIP is reached. This approach enables LVLM to focus only on processing a few high-resolution image tiles in a coarse-to-fine manner, thereby reducing computational complexity while preserving critical text-related image details.

\begin{algorithm}[t]
\caption{Coarse-to-Fine Token Pruning with DIP}
\label{alg:pyramid_construction}
\begin{algorithmic}[1]

\REQUIRE $ \{I^1_\text{tiles}, I^2_\text{tiles}, \dots, I^{\text{P}}_\text{tiles}\}$ in $I_{\text{DIP}}$, $K$-layer RFM, initial multimodal tokens $T$, vision encoder and projector $\mathcal{V}(\cdot)$, token saving ratio $\alpha$, tile number threshold $N_{\max}$
\ENSURE Retained tokens $T^{\text{retain}}_{\text{vis}}$

\STATE \textbf{Definition:} $\text{Top-}\alpha(X)$ selects elements in $X$ with values in the top $\alpha$ proportion.

\FOR{$p = 1$ \TO\ $P$}
    \STATE Compute attention: 
      $\smash{A^{K,hr}_{\text{stu}}} \gets \text{RFM}(T)[\smash{T^{p,hr}_{\text{vis}}}]$\vphantom{$\textstyle A^{K}_{N}$}
    \STATE Normalize: 
     \STATE Identify key positions: 
     $\smash{\mathcal{P}^{K,hr}_{\text{key}}} \gets \text{Top-}\alpha(\smash{A^{K,hr}_{\text{stu}}})$
    \IF{$p < P$}
        \STATE Map coordinates and select key tiles: 
          $I^{p+1}_\text{key} \gets \{t \in I^{p+1}_{\text{tiles}} \mid \text{mapped from } \mathcal{P}^{K,hr}_{\text{key}}\}$\vphantom{$\textstyle \mathcal{S}^{P}_{\text{key}}$}
    
        \IF{$|I^{p+1}_\text{key}| > N_{\max}$}
             \STATE Prune tokens: 
               $T^{\text{retain}}_{\text{vis}} \gets \{t \in T^{p,hr}_{\text{vis}} \mid \smash{\mathcal{P}^{K,hr}_{\text{key}}}\}$ 
             \STATE \textbf{break}
        \ELSE
            \STATE Encode selected tiles: 
              $\smash{T^{p+1,hr}_{\text{vis}}} \gets \mathcal{V}(I^{p+1}_\text{key})$
            \STATE Update $T \gets [\smash{T^{\text{lr}}_{\text{vis}}}, \smash{T^{p+1,hr}_{\text{vis}}}, T_{\text{txt}}]$\vphantom{$\textstyle A^{K}_{N}$}
        \ENDIF
    \ELSE
        \STATE Prune tokens at final layer: 
        $T^{\text{retain}}_{\text{vis}} \gets \{t \in T^{p,hr}_{\text{vis}} \mid \smash{\mathcal{P}^{K,hr}_{\text{key}}}\}$
    \ENDIF
\ENDFOR

\RETURN $T^{\text{retain}}_{\text{vis}}$
\end{algorithmic}
\end{algorithm}

\begin{table*}[!t]
\centering
\renewcommand{\arraystretch}{1.2} 
\begin{tabular}{cccccc}
\toprule
Dataset       & Image Size  & Image Num & Ques. Cate. & Ques. Num & QS Format \\ \hline
MME-RealWorld-RS~\cite{zhang2024mme}   & 689-11500   & 1298  & 3 & 3738 & Single-choice \\
LRS-VQA (Ours)   & 1024-27328 & \textbf{1657}     & \textbf{8}     & \textbf{7333} &  Open-end \\
\bottomrule
\end{tabular}
\caption{Comparison of existing benchmarks for evaluating LVLMs' perception capabilities in large RSIs.}
\label{tab:Data_compare}
\vspace{-5pt}
\end{table*}

\subsection{The Construction of LRS-VQA Benchmark}

Considering the limitations of existing benchmarks for evaluating LVLMs' perception of RSIs in terms of question diversity and image size, we propose a new benchmark for more comprehensive evaluation.

\begin{figure}[!h]
    \centering
    \includegraphics[width=\columnwidth]{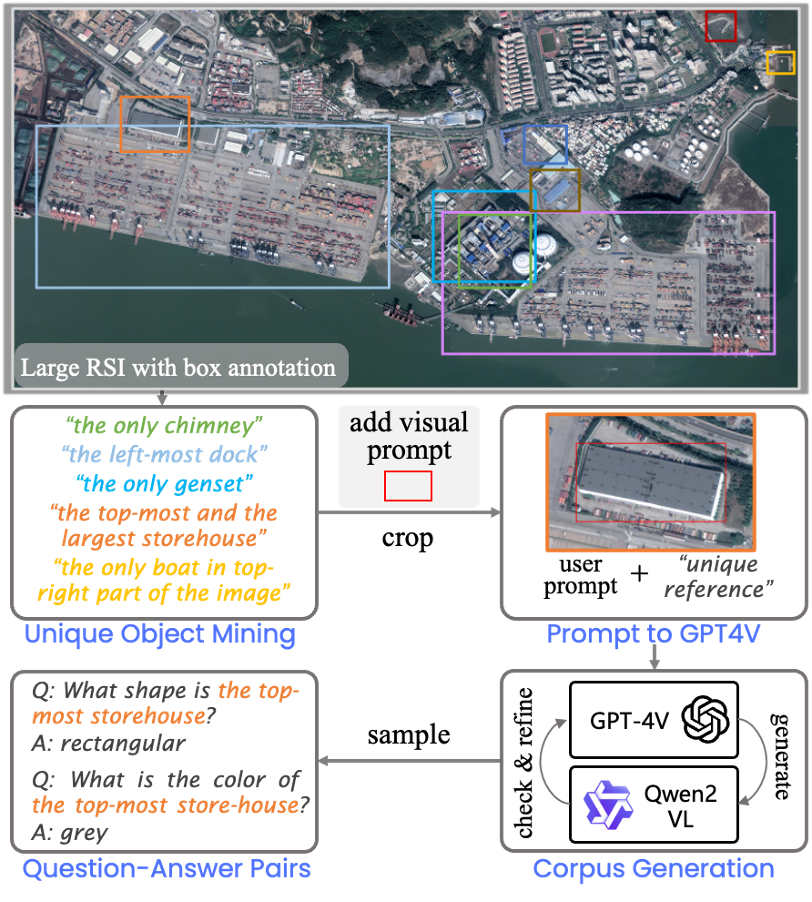}
    \vspace{-14pt}
    \caption{The construction pipeline of the proposed LRS-VQA dataset. The visual prompt (red box) is inspired by SoM~\cite{yang2023setofmark}.}
    \vspace{-10pt}
    \label{fig:vqa_anno_pipeline}
\end{figure}

\subsubsection{Annotation Pipeline}

The annotation pipeline of LRS-VQA is illustrated in Fig.~\ref{fig:vqa_anno_pipeline}. We collect 3 publicly remote sensing datasets: FAIR1M-1.0~\cite{sun2022fair1m}, GLH-Bridge~\cite{li2024learning}, and STAR~\cite{li2024scene}, and identify all unique targets based on the object detection labels to generate unique references from them (e.g., ``the top-most ship" and ``the largest dock"). Subsequently, we crop the region with the unique target from the original image, and feed it, along with a prompt containing the unique reference, into GPT-4V~\cite{gpt4v} to generate question-answer pairs about the target's color, shape, status, etc. Details of the overall construction process are provided in the Appendix~\ref{appendix_a}.

\subsubsection{Visual-Centric Quality Check and Refinement}

\begin{figure}[!t]
    \centering
    \includegraphics[width=\columnwidth]{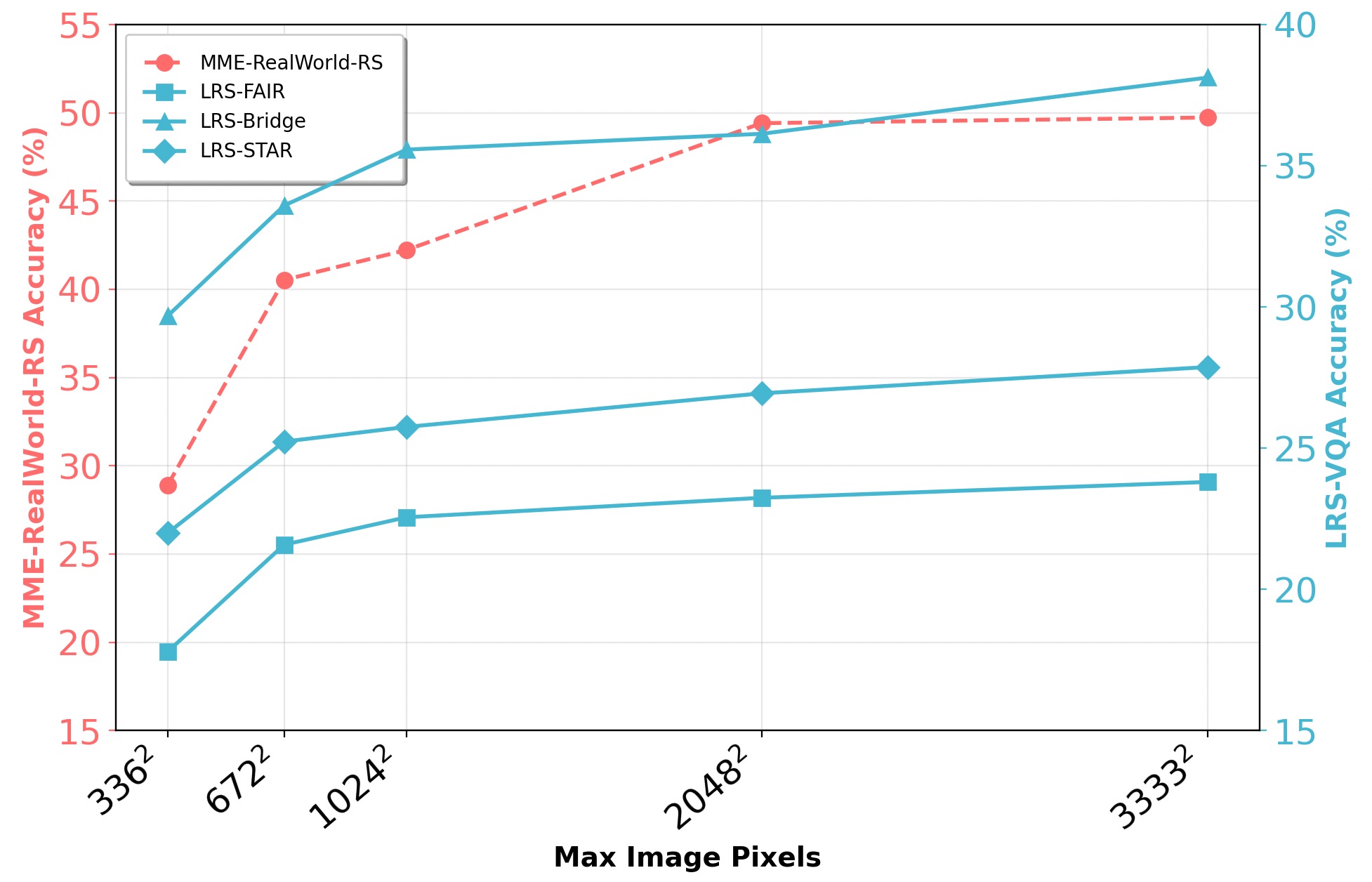}
    \caption{The accuracy trends of Qwen2-VL across varying input maximum pixels. This demonstrates that accuracy on both the manually annotated MME-RealWorld-RS and our proposed LRS-VQA exhibit a positive correlation with resolution improvement, proving the effectiveness of LRS-VQA in evaluating LVLM's high-resolution RSI perception capabilities. }
    \label{fig:acc_line}
    \vspace{-5pt}
\end{figure}

To ensure the quality of the generated question-answer pairs, we employ powerful Qwen2-VL~\cite{wang2024qwen2vl} for quality assessment. Qwen2-VL supports manual adjustment of the maximum pixel resolution, allowing us to evaluate whether increased resolution improves accuracy while keeping the LLM unchanged. Using this approach, we filter out question types where accuracy can not improve with resolution. Through iterative filtering, manual review, and prompt refinement, we obtain the LRS-VQA dataset, which could effectively assess the large RSI perception capability of LVLMs, as shown in Fig.~\ref{fig:acc_line}. LRS-VQA contains eight question-answer types: \textit{count, color, category, shape, status, reasoning, rural/urban classification}, and \textit{target background}. It features greater diversity in question types and larger image sizes than MME-Realworld-RS as in Tab.~\ref{tab:Data_compare}, highlighting the complexities of large RSI perception.

\section{Experiments}

\begin{table*}[!h]
\vspace{-2pt}
\renewcommand{\arraystretch}{0.8}
\small
\centering
\renewcommand{\arraystretch}{0.95} 
\setlength{\tabcolsep}{5pt} 
\begin{tabular}{l | c c c c | c c c c c}
\toprule
\textbf{Leaderboard} &  \textbf{Data}  & \textbf{Vis. Encoder}    & \textbf{LLM}     & \textbf{Max Size} & \textbf{\makecell{MME-\\RW-RS}} & \textbf{\makecell{LRS-\\FAIR}} & \textbf{\makecell{LRS-\\Bridge}} & \textbf{\makecell{LRS-\\STAR}}  & \textbf{\makecell{Avg.\\Acc}} \\ \hline
Qwen2-VL~\cite{wang2024qwen2vl} & - & \makecell{QwenViT} & Qwen2-7B & 3,333×3,333 & 49.73  & 23.80  & 38.12  & 27.87 &  34.88 \\
LLaVA-OV~\cite{li2024llava} & 4.8M & SigLip & Qwen2-7B & 2,304×2,304 & 53.53  & 20.61  & 35.11  & 26.08 &  33.83 \\
IXC-2.5~\cite{internlmxcomposer2_5} & -  &  CLIP-L14   &  InternLM2-7B                   &   4,096×4,096   &  36.12 & 25.25 & 38.41 & 27.30 & 31.77\\ 
LLaVA-UHD-v2~\cite{zhang2024llavauhdv2} & 1.42M &  \makecell{CLIP-L14+JBU} & Vicuna-1.5-7B & 672×1,008 & 40.77  & 22.82 & 32.57 & 26.08 & 30.56 \\ 
LLaVA-FlexAttn~\cite{li2024flexattention} & 1.22M &CLIP-L14 & Vicuna-1.5-7B  & 1,008×1,008 & 37.75 & 19.57 & 29.99 & 22.76  &  27.52   \\ 
SEAL~\cite{wu2024vstar} & 0.95M & CLIP-L14 & Vicuna-1.5-7B & - & 30.55 & 21.29 & 34.75 & 21.29 & 26.97 \\
 MGM-HD~\cite{li2024mgm} & 3.0M & Mixed & Vicuna-1.5-7B &  1,536×1,536  & 31.51 & 17.90 & 35.92 & 20.13 & 26.36 \\
 SliME~\cite{zhang2024slime} & 2.0M & CLIP-L14 & Vicuna-1.5-7B & 672×1,008 & 28.54 & 17.11 & 32.09 & 22.99 & 25.18 \\
LLaVA-1.5~\cite{liu2023improved} & 1.22M & CLIP-L14 & Vicuna-1.5-7B & 336×336 & 26.38 &18.76 & 30.70 & 22.63 & 24.62 \\  \hline
  RSUniVLM~\cite{liu2024rsunivlm}  &   1.2M  & SigLIP  &  Qwen2-0.5B & 336×336  &   25.39  & 21.02 	 & 32.61  & 24.72  & 25.93  \\
 Geochat~\cite{kuckreja2023geochat} & 1.53M & CLIP-L14 & Vicuna-1.5-7B & 504×504 & 28.62 & 20.18  & 24.54 & 13.75  &  21.77 \\  \hline
  GPT-4o~\cite{gpt4o} & - & - & - & - & 28.92 & 22.15 & 31.84 & 27.40 & 27.58 \\
   GPT-4o-mini~\cite{gpt4o} & - & - & - & - & 6.69 & 18.67 & 31.99 & 25.85 & 20.80\\
  Claude-3.5-Sonnet~\cite{claude35} & - & - & -& - & 25.74 & 12.95 & 26.69 & 13.29 & 19.67 \\
\hline
 \midrule  
 \textbf{Comparison} & &  &   &  &  &   &   &  \\ \hline
 \textit{LLaVA-1.5*} &\multirow{6}{*}{1.04M} & CLIP-L14 & \multirow{6}{*}{Vicuna-1.5-7B} & 336×336 & 34.40 &  18.24 & 32.28 & 24.17  & 27.30  \\
\textit{LLaVA-1.5-p4*} & & CLIP-L14 &  & 672×672 &38.20 & 19.18 & 35.43 & \underline{26.50} & 29.83 \\
\textit{Geochat*} & & CLIP-L14 &   & 504×504 & 33.68 & 21.51   & 35.97 &  25.86  &  29.25  \\ 
\textit{SLiME*} & & CLIP-L14 &  & 672×1,008 & 34.56 & \underline{21.98} & 33.20 & 25.10 & 28.71 \\
\textit{LLaVA-UHD-v2*} & & CLIP-L14+JBU &   & 672×1,008 & \underline{38.55} & 20.77 & \underline{36.57} & 25.79 & \underline{30.42} \\
\rowcolor{gray!20} {Ours (LLaVA-1.5)} & & CLIP-L14 &  & Dynamic & \textbf{39.04} & \textbf{22.97} & \textbf{36.89} & \textbf{27.48} & \textbf{31.59} \\ \hline
\textit{LLaVA-Next-p4*} &  \multirow{4}{*}{1.04M}  & \multirow{4}{*}{CLIP-L14} &  \multirow{4}{*}{Qwen2-7B} & 672×672 & 39.12 &  21.06  & 36.27 & 25.80 & 30.56  \\ 
\textit{LLaVA-Next-p9*} &  &  &  & 1,008×1,008  &  \underline{40.35} & \underline{21.14} & \underline{37.25} & 26.10 & \underline{31.21} \\ 
\textit{LLaVA-Next-p25*} &  &  &   & 1,680×1,680 & 39.65 & 20.99 & 36.38  & \underline{26.18}  & 30.80  \\  	 	
\rowcolor{gray!20} {Ours (LLaVA-Next)} & &   &   & Dynamic & \textbf{41.89} & \textbf{21.85}  & \textbf{38.24}  & \textbf{26.67} & \textbf{32.16} \\ \hline  		 	
\bottomrule
\end{tabular}
\vspace{-3pt}
\caption{Leaderboard and performance comparison on LRS-VQA and MME-RealWorld-RS. ``MME-RW-RS" indicates the MME-RealWorld-RS. ``Data" refers to the total instruction data used during PT and SFT. ``\textit{p}$X$" indicates the max number is $X$ for the pre-defined grids. ``\textit{method*}'' indicates we reproduce the SFT stage of existing methods using the 484k instruction data.}
\label{tab:main_res}
\vspace{-5pt}
\end{table*}

\begin{table}[htbp]
\centering
\small
\renewcommand{\arraystretch}{0.98} 
\setlength{\tabcolsep}{2.8pt} 
\begin{tabular}{lcccccccc}
\toprule
\textbf{Setting} & \textbf{Color} & \textbf{Count} & \textbf{Position} & \textbf{Acc} & \textbf{FPS}   \\
\midrule
anyres-p25 &   41.56    & 31.05  & 46.14   & 39.65  & 0.188 \\
\textit{w/ PruMerge++}~\cite{shang2024llavaprumerge} &  43.27 & 28.38 & 32.78 & 34.86 &  0.152 \\
\textit{w/ VisionZip}~\cite{yang2024visionzip}  & 42.71 & 24.55 & 34.55& 33.98 & 0.183 \\ 
\textit{w/ FastV}~\cite{chen2024fastv}  & 39.52 & 30.10 & 43.99 & 37.93 & \underline{0.192}\\
\textit{w/ PDrop}~\cite{xing2024pyramiddrop}  & 41.00 & 30.54 & 47.77 & 39.85 & 0.184 \\
\textit{w/ prune (CLIP)}    & 37.80  & 27.65  & 44.90 & 36.86 & 0.171 \\
\textit{w/ prune (RemoteCLIP)}  & 39.40 & 29.83 & 47.45  & 38.77 & 0.148 \\
\rowcolor{gray!20}\textit{w/ prune (Ours)} & 43.98 & 30.42 & 49.16 & \underline{41.28} & 0.165 \\ \hline
 \makecell[l]{\text{DIP-3layer}\\ \textit{w/ prune (Ours)}}  & 44.08 & 30.94 & 49.26 & \textbf{41.31} & \textbf{0.267}  \\
\bottomrule
\end{tabular}
\vspace{-3pt}
\caption{Comparison with different token reduction methods based on LLaVA-Next-Qwen2. The ``anyres-p25'' setting means the maximum number of pre-defined grids is 25 (1680×1680 pixels). FPS is computed based on the inference time for continuous 800 images from MME-Realworld-RS.}
\label{tab:ablation_token_reduce}
\vspace{-12pt}
\end{table}

\begin{table}[htbp]
\centering
\small
\setlength{\tabcolsep}{4pt} 
\renewcommand{\arraystretch}{0.98} 
\begin{tabular}{lcccc}
\toprule
\textbf{Setting} & \textbf{\makecell{Vis. Tokens\\(Total)}} & \textbf{\makecell{Vis. Tokens\\(to LLM)}} & \textbf{\makecell{TFLOPs\\(B)}} \\
\midrule
anyres-p144 & 83520 & 21312 & 243.37  \\
\textit{w/ PruMerge++}~\cite{shang2024llavaprumerge} & 83520 & \underline{5328} & \underline{43.75} \\
\textit{w/ VisionZip}~\cite{yang2024visionzip}  & 83520 & 9280 & 83.56 \\
\textit{w/ FastV}~\cite{chen2024fastv}  & 83520 & 21312 & 109.21 \\
\textit{w/ PDrop}~\cite{xing2024pyramiddrop} & 83520 & 21312  & 101.62 \\
\rowcolor{gray!20}\textit{w/ prune (Ours)} & 83520 & 5760 & 82.56 \\ \hline
\makecell[l]{\text{DIP-4layer}\\ \textit{w/ prune (Ours)}}  &  \textbf{55296} & \textbf{2376}  & \textbf{36.61} \\
\bottomrule
\end{tabular}
\vspace{-3pt}
\caption{Comparison of computational efficiency with different token reduction methods based on LLaVA-Next-Qwen2. We assume inferring a 4000×4000 pixel image (anyres-p144) and report the theoretical TFLOPs of vision tokens after the vision projector.}
\label{tab:ablation_flops_test}
\vspace{-14pt}
\end{table}

\subsection{Experimental Details}

\textbf{Data.} Our pre-training (PT) phase is the same as LLaVA-1.5~\cite{liu2024visual} (using 558K data). For the SFT phase, we collect 484K samples, including 300K sampled from LLaVA-1.5-665K, 146K sampled from RSVQA-HR~\cite{lobry2020rsvqa}, and 38K template-based samples from three RS datasets~\cite{ding2021object,li2024learning,li2024scene} using their labels. Within this 38k data, some images share the same source with LRS-VQA, but there is no overlap between them. Details can be found in the Appendix~\ref{appendix_b1}.

\textbf{Benchmarks and evaluation metrics:} i) MME-RealWorld-RS: the RS part of MME-Realworld~\cite{zhang2024mme}, containing 1,298 RSIs with expert-annotated questions in three types: color, count, and position. We follow the official evaluation script but modify the prompt by removing ``The best answer is:" to address Vicuna-1.5-based models' tendency to respond with only option A. ii) LRS-VQA: it consists of 3 parts: LRS-FAIR, LRS-Bridge, and LRS-STAR, containing 2,272, 1,062, and 3,999 QA pairs, respectively. For the short open-ended format, we adopt a structured evaluation metric following~\cite{hudson2019gqa,xiao2021next}, using WordNet~\cite{miller1995wordnet}, with a semantic similarity threshold of 0.8.

\textbf{Experimental settings.} We evaluate existing LVLMs using their official weights or APIs. For fair comparison, we also re-implement some methods on our 484k SFT dataset. Our approach is built upon the LLaVA-1.5 codebase, where we integrate the anyres strategy and Qwen2 language model. All experiments are conducted on 4 A100 80GB GPUs. For our method, the minimum length for DIP is 1,008 pixels. The $N_{\max}$ is set to 40 and 80 for Vicuna-1.5 and Qwen2, respectively. The vision token saving ratio $\alpha$ is 0.25, $\lambda_{\text{hr}},\lambda_{\text{mse}}, \lambda_{\text{kl}}$ are set to 2.0, 1.0, 1.0. The RFM-LLM layer pairs are $[1,5,11,14]$ with distillation applied to the first and last pairs. Additional experimental details are provided in Appendix~\ref{appendix_b2}.

\subsection{Leaderboard and Comparison}

\begin{table}[tbp]
\centering
\small
\setlength{\tabcolsep}{4pt} 
\begin{tabular}{l | cccccc}
\toprule
\textbf{Method} & \textbf{\makecell{Drop\\ Ratio}}  & \textbf{Color} & \textbf{Count}  & \textbf{Pos} & \textbf{Acc} \\
\midrule
anyres-p25 & 0\% & 41.56   & 31.05  & 46.14   & 39.65 \\
\multirow{4}{*}{\textit{w/ prune (Ours)}} & 25\% & 42.47 & 31.08 & 47.41 & 40.40 \\
  & 50\% & 42.39 & \textbf{31.24} & 48.45 & 40.77 \\
 & 75\% & \textbf{43.98} & 30.42 & \textbf{49.16} & \textbf{41.28}\\
  & 90\% & 41.51 & 29.77 & 46.70 & 39.41 \\
\bottomrule
\end{tabular}
\vspace{-3pt}
\caption{Ablation study on different prune ratios with LLaVA-Next-Qwen2, when pruning vision tokens based on RFM results without DIP. This table demonstrates that pruning irrelevant tokens in high-resolution RSIs can improve performance.}
\label{tab:prune_ratios}
\vspace{-2pt}
\end{table}

\begin{table}[tbp]
\centering
\small
\begin{tabular}{cccccc}
\toprule
\textbf{Max Size} &  \textbf{Color} & \textbf{Count} & \textbf{Pos} & \textbf{Acc}  & \textbf{FPS}\\
\midrule
1,008×1,008 & 42.87 & 29.85 & 47.89 & 40.29 & \textbf{0.271} \\
2,016×2,016 & 44.08 & \underline{30.94} & \underline{49.26} & \underline{41.31} & 0.267\\
4,032×4,032 & \textbf{44.86}& 29.85 & 47.18 & 40.72 & 0.221 \\ 
8,064×8,064 & 43.98  & 30.42 & 45.11 & 39.91  & 0.204\\
\hline
Dynamic & \underline{44.70} & \textbf{31.00} & \textbf{49.72} & \textbf{41.89} & \underline{0.238}\\
\bottomrule
\end{tabular}
\vspace{-3pt}
\caption{Ablation study on fixing the number of DIP layers reached in inference with LLaVA-Next-Qwen2. Rows 1-4 correspond to fixing the number of DIP layers being 2-5, respectively.}
\label{tab:ablation_pyramid_layers}
\vspace{-2pt}
\end{table}

\textbf{Leaderboard on LRS-VQA.} We evaluate a total of 11 open-source LVLMs, including methods for high-resolution ~\cite{liu2024visual,zhang2024slime,internlmxcomposer2_5,li2024flexattention}, chain-of-thought reasoning~\cite{wu2024vstar}, multiple visual encoders~\cite{li2024mgm}, and RS LVLMs~\cite{kuckreja2023geochat,liu2024rsunivlm}. We also evaluate 3 closed-source MLLMs. Results in Tab.~\ref{tab:main_res} indicate that LRS-VQA is more challenging than MME-RealWorld-RS on average, reflecting the complexities of large RSIs.

\textbf{Comparison with high-resolution methods.} As shown in Tab.~\ref{tab:main_res}, we select 5 different high resolution strategies, performing SFT on the same dataset for comparison. Our method performs better across all 4 datasets.

\textbf{Comparison with token reduction methods.} We reproduce 4 plug-and-play methods and establish 2 key baselines for comparison on the MME-RealWorld-RS. These baselines utilize sliding-window similarity maps generated by CLIP-L14~\cite{radford2021learning} and RemoteCLIP-L14~\cite{liu2024remoteclip}, respectively. More details are in the Appendix~\ref{appendix_b2}. As shown in Tab.~\ref{tab:ablation_token_reduce}, our method achieves the best performance under the fixed anyres-p25 resolution. It also secures the highest accuracy and FPS in the 3-layer DIP setting (up to 2,016×2,016 pixels). For Tab.~\ref{tab:ablation_flops_test}, we report the average number of tiles selected by our dynamic method for images of similar sizes. These results suggest that existing methods often incur high computational costs when processing numerous text-irrelevant image tiles in large images.

\subsection{Ablation Results}

\textbf{Different ratios for RFM-based pruning.} 
As shown in Tab.~\ref{tab:prune_ratios}, a key finding is that in large RSIs, the complex background and the small foreground regions enable high pruning rates to deliver performance benefits.

\textbf{Fixed DIP layers for inference.} 
As shown in Tab.~\ref{tab:ablation_pyramid_layers}, we fix the number of DIP layers reached during inference. If an image's DIP is shallower than the specified number, the last layer is used. Results show that forcing traversal to higher resolutions can degrade performance since not all questions require image details. Therefore, our strategy dynamically selects the termination DIP layer based on the input text, balancing accuracy and efficiency.

\begin{table}[tbp]
\centering
\small
\setlength{\tabcolsep}{4pt} 
\begin{tabular}{l | c c c}
\toprule
\textbf{Pruning Guidance} & \textbf{LRS-VQA} & \textbf{DIOR-RSVG} & \textbf{RRSIS-D} \\ 
\midrule
Teacher LLM Attn.  &  56.14   &  74.81   & 73.88                \\ \hline
CLIP Sim.        &   20.14   &     29.10  &   27.53             \\
RemoteCLIP Sim.    &  30.41  &  43.12  &    41.81         \\
RFM Attn. (Ours)      &   \textbf{47.99}  &   \textbf{64.76}   &  \textbf{61.17}   \\  
\bottomrule
\end{tabular}
\vspace{-5pt}
\caption{Localization recall (\%) of different pruning guidance.}
\label{tab:location_acc}
\vspace{-2pt}
\end{table}

\begin{figure}[!t]
    \centering
    \includegraphics[width=\columnwidth]{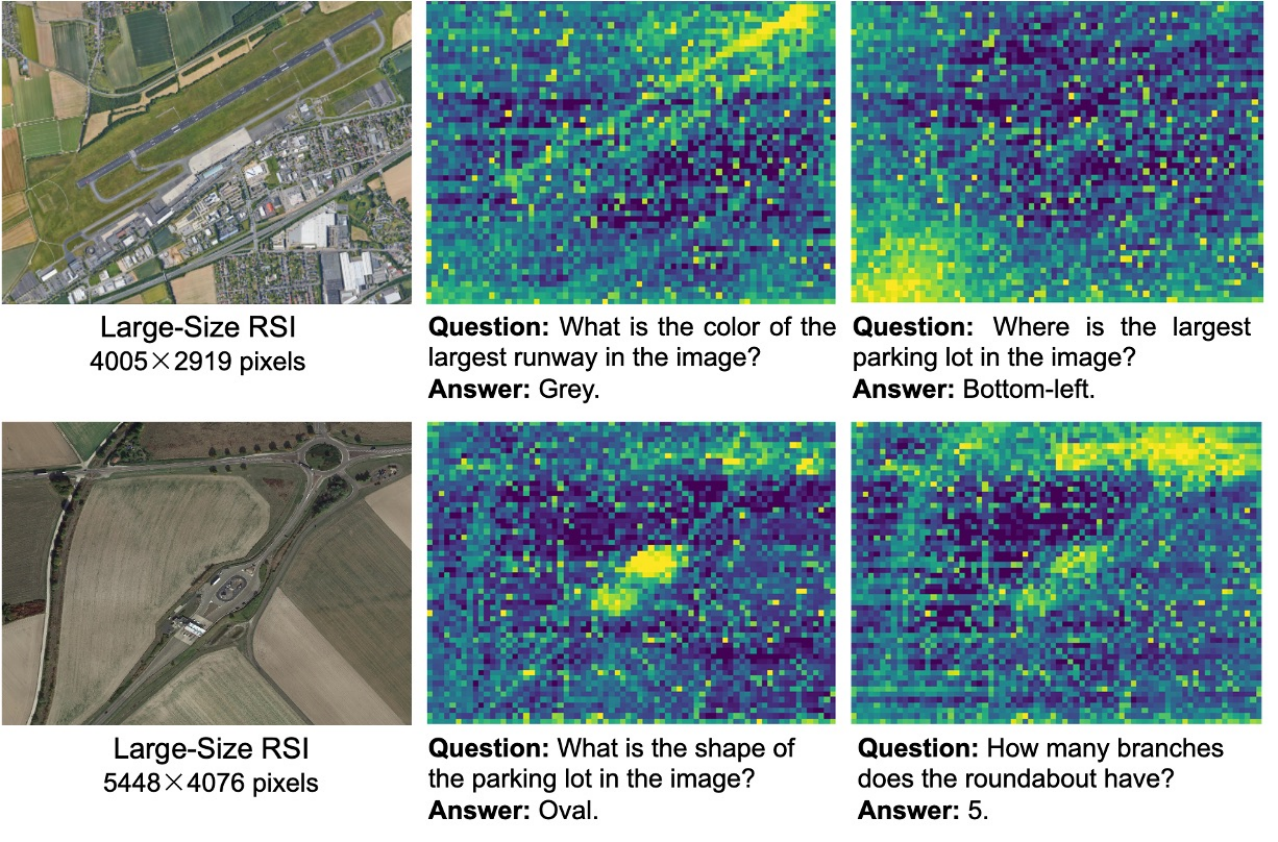}
    \vspace{-14pt}
    \caption{Text-related visual attention localization in the last layer if trained RFM under LLaVA-Next-Qwen2.}
    \label{fig:ref_vis1_mask}
    \vspace{-8pt}
\end{figure}

\textbf{Attention Localization validation.} We select LRS-VQA (using boxes of cropped regions) and two referring-based datasets, DIOR-RSVG~\cite{zhan2023rsvg} and RRSIS-D~\cite{liu2024rotated} for localization accuracy evaluation, using a \textbf{recall metric}: a prediction is considered successful if the vision tokens retained after pruning cover more than 50\% of the ground-truth region. The results in Tab.~\ref{tab:location_acc} demonstrate that at a 25\% token retention ratio, both the teacher LLM's attention and our student RFM achieve higher localization accuracy than the CLIP-L14 and RemoteCLIP-L14 baselines (also based on sliding-window similarity maps). This confirms that LLM attention provides a reliable localization signal and our distillation is effective.

\textbf{Visualization results.} In Fig.~\ref{fig:ref_vis1_mask}, we visualize the attention maps generated by the trained RFM for vision tokens from initial input image tiles. The results demonstrate that text references to targets effectively guide the RFM's attention output, supporting our coarse-to-fine inference.

 \vspace{-5pt}
\section{Conclusion}
\vspace{-4pt}
This paper presents a text-guided token pruning method tailored for efficiently processing large remote sensing images (RSIs). To address the limitations of existing high-resolution approaches when handling large RSIs, our method integrates the Region Focus Module (RFM) and Dynamic Image Pyramid (DIP) to focus on text-relevant key vision tokens while reducing computational overhead. Furthermore, we introduce a new benchmark, LRS-VQA, with 7,333 QA pairs covering 8 question types, and features larger image sizes and a diverse range of question types. 

\textbf{Future work.} Vision-language understanding of large RSIs remains a challenging task. Investigating approaches such as long-context transfer from LLMs to LVLMs, or chain-of-thought reasoning, is also meaningful.

\section*{Acknowledgment}

The authors would like to thank the anonymous reviewers for their valuable comments. This work was supported by the National Natural Science Foundation of China under Grants 42371321, 424B2006, and the Natural Science Foundation of Shanghai under Grant No. 25ZR1402268.


{
    \small
    \bibliographystyle{ieeenat_fullname}
    \bibliography{main}
}


\clearpage

\begin{appendices}
\setcounter{table}{0} 
\setcounter{figure}{0} 
\setcounter{equation}{0} 

\setcounter{page}{1}
\maketitlesupplementary

\section{Detailed Construction Process of LRS-VQA}
\label{appendix_a}

In this section, we describe the construction process of LRS-VQA in detail, including data collection and filtering, label creation, and quality assurance.

\subsection{Unique Object Extraction}

Given the challenge of precisely referring to a specific unique object in large Remote Sensing Images (RSIs) (e.g., identifying a particular ship among hundreds of ships parking on the same harbor), we perform rule-based unique object extraction using object detection labels from our collected RS datasets~\cite{sun2022fair1m,li2024learning,li2024scene}. The process is detailed below:

(i) We first filter out small images and calculate the total number of instances for each category in the image and remove categories with more than 40 instances per image. 

(ii) For the remaining categories, we extract attributes like \textit{absolute position}, \textit{absolute size}, \textit{relative position} and \textit{relative size} within the same category. Based on this information, we determine whether an object is unique and create \textbf{unique reference} (e.g., ``the top-most airplane" or ``the only storehouse in the bottom-left corner of the image"). Note that multiple thresholds are set during this process. For example, a target is labeled as the distinguishable ``largest" only if its area exceeds that of the second-largest target in the same category by more than 20$\%$. Similarly, a target is marked as ``right-most" within its category only if it is located farthest to the right and its offset distance from the next closest target is greater than 20 pixels.

(iii) Based on the above results, we crop the region containing the unique targets from the large RSI and draw a red box around the object as the visual prompt. For small targets, if the longer side of the target is less than 400 pixels, the cropping area is expanded by 400 pixels. For larger targets, we apply appropriate scaling to ensure that the longer side does not exceed 1400 pixels.

Finally, for each large RSI, we obtain local image patches containing unique targets, along with their corresponding unique references.

\subsection{Question-Answer Pair Generation}

\begin{table*}[h!]\centering
 \vspace{10pt}
\begin{minipage}{0.99\linewidth}\vspace{0mm}    \centering
\begin{tcolorbox} 
    \centering
     \hspace{-6mm}
    \begin{tabular}{p{0.99\linewidth}}
\begin{minipage}{0.99\linewidth}\vspace{0mm}

\VarSty{messages} = [
            \{\var{"role":"system", "content":} f``````You are an AI visual assistant tasked with analyzing remote sensing images. Given the visual input (a part of a large image) and corresponding object information, your job is to create a list of question-answer pairs around the target object and its surroundings. Each sentence should unambiguously refer to the object based on the `why unique' information.  \\
            Finally, you need to return [`qa-pairs': [`ques-id': question id, `question': question, `type': question type, `answer': answer]] in JSON format. Do not return any notes after the JSON. The target object is highlighted by a red rectangle in the given image patch, and the `why-unique' provides how to refer it in the original large image, you need to rely on this information to ask questions. \\
            
            1. Based on all visible elements and object information, ask 5-10 questions of various types, including object existence, object relation, complex reasoning, and object status. Avoid questions about color, object shape and object orientation. Additionally, questions requiring reasoning should involve multifaceted analytical thought processes (e.g., analyzing object distribution patterns) based on the target object and its surroundings. Possibly include objects that are not provided, such as houses, roads, water and trees if they are obvious and non-ambiguous. \\
            2. Ensure each question has a definite answer without any ambiguity, and answer each question using a single word or phrase, no more than 3 words.\\
            3. Only ask questions about clear answers; avoid uncertainties or unclear elements, such as unknown, uncertain, some, or several. \\
            4.Avoid question formats that only allow for two options or overly simplistic responses (e.g., `Yes' or `No'). \\
            5. Do not mention the red highlight box or asking about the target object's category—consider it known.\\
            6. Use complete information from `why-unique' to ensure unique reference. 
            
           Follow the above guidelines and ensure consistency with the provided category."""\}\\
        ]
    \var{\VarSty{messages}.append(\{"role":"user", "content":`\textbackslash  n'.join(\VarSty{query})\})}
\end{minipage}
    \end{tabular}
\end{tcolorbox}

\vspace{-2mm}
 \caption{The prompt to GPT-4V for generating question-answer pairs about the unique objects in the large-size RSIs.}
\label{tab:gpt4v-prompt}
\end{minipage}
\end{table*}

Based on the above information, we additionally filter out extremely small objects (smaller than 16×16 pixels). Then we design prompts as in Tab.~\ref{tab:gpt4v-prompt} to instruct GPT-4V to generate a diverse set of question-answer pairs. We carefully design the prompt to avoid generating questions about the entire image (e.g., counting targets across the whole image). Questions involving whole-image counting for specific categories are separately generated based on object detection labels.

During the initial generation of the VQA corpus, we observed that the answers to VQA questions were predominantly ``yes" or ``no". This could lead to the LVLM achieving high accuracy even without visual input. To address this issue, we carefully refine the prompts to guide GPT-4V in generating diverse and informative responses, constrained to a length of 1 to 3 words, while ensuring the responses are provided in an open-ended format. For the final version filtered by Qwen2-VL, we then conduct expert spot-checking and correction.

Additionally, during the Qwen2-VL-based quality inspection process and manual check, we observed that GPT-4V exhibits limitations in handling certain types of questions specific to remote sensing scenarios, such as object orientation. This indicates that even state-of-the-art LVLMs still need improvement when interpreting RSI.

\section{More Experiment Details}
\label{appendix_b}
\subsection{Training Data Construction}
\label{appendix_b1}

We first filter out excessively large images from three RS datasets, as the DIP has a fixed 2-layer structure during training.  To ensure the GSD range during training covers the dynamic range in inference, we apply different downsampling factors to maintain varied image sizes and construct multi-GSD inputs.

Subsequently, we create two types of questions: count-type questions based on object detection labels and relation-based questions using scene graph annotations. For count-type questions, we avoid querying categories with excessively high counts and control the proportion of samples where the answer is ``1". For relation-based questions, we ask whether a specific relationship exists between any two categories in the image, with answers provided as ``yes" or ``no".

\subsection{Implementation Details}
\label{appendix_b2}
\textbf{Training setting.} During training, we follow the same experimental setup of the SFT stage as LLaVA-1.5 and LLaVA-Next, with a global batch size of 128 and a learning rate of 2e-5. Additionally, the maximum length of Vicuna-1.5 and Qwen2 are 2048 and 16384, respectively. The overall SFT process is largely consistent with that of LLaVA-1.5, as our RFM is a plug-and-play module that can be seamlessly integrated into the SFT process of any modular LVLM.

\textbf{More details about coarse-to-fine token pruning.} Our method involves stopping at a specific layer $p$ of the DIP during traversal, pruning based on the RFM output of that layer, and then concatenating the pruned vision tokens with the vision tokens from the thumbnail view, along with text instructions, to form the multimodal input for the LLM. Although only a subset of image tiles is selected when traversing the DIP, we pad the unselected image tiles after the vision encoder and then add the ``image newline" delimiter along with global position embeddings. Subsequently, we extract the vision tokens corresponding to $I^{p+1}_\text{key}$ and the ``image newline" positions from the fully padded tensor, which serve as $\smash{T^{p+1, hr}_{\text{vis}}}$. The definitions of these symbols are consistent with those in the main paper.

\textbf{Maximum sequence length of LLM.} For Vicuna-1.5, due to the limitations of its pre-trained weights, it is challenging to train its long-context processing capability from scratch. Additionally, its performance often degrades when extrapolating to longer sequences. As for Qwen2, we have not yet explored the impact of extending the maximum sequence length to cover all vision tokens from original resolution imagery, primarily due to the significant time and resource costs. Moreover, our method does not rely on enhancing long-sequence processing capabilities to handle large images but rather serves as a strategy to improve the perception performance of existing modular LVLMs.

\textbf{More details of comparison methods.} 

During the comparison of token reduction methods, we strive to maintain consistency or make necessary adaptations due to the differing principles of each approach. Specifically, for PruneMerge++~\cite{shang2024llavaprumerge}, CLIP-based pruning, RemoteCLIP-based pruning, and our RFM-based pruning, we set the token retain ratio to 25$\%$. For VisionZip~\cite{yang2024visionzip}, the number of retained tokens is set to 64. For PyramidDrop~\cite{xing2024pyramiddrop} under the Qwen2 backbone, the pruning layers are configured to [7, 14, 21]. For all other settings of FastV~\cite{chen2024fastv} and PyramidDrop, we adhere to their original implementations.

\subsection{Flash Attention and Multi-Turn Support}
\label{appendix_b3}
Our method follows existing token pruning approaches~\cite{zhang2024sparsevlm,xing2024pyramiddrop} and is compatible with the use of flash attention as well as multi-turn dialogue 

For flash attention, following SparseVLM~\cite{zhang2024sparsevlm}, we introduce an additional forward pass that incorporates a specially designed value matrix. This allows us to extract the mean value of the processed attention map without explicitly computing the full attention map. 

For multi-turn conversations, our method offers advantages over existing grid-based dynamic high-resolution approaches. These methods typically rely on pre-defined grids to partition and store all corresponding image tiles, which can be memory-intensive. In contrast, our approach only caches the features from the first two layers of the DIP (i.e., the vision tokens from the thumbnail view and the first group of image tiles). For higher-resolution image tiles, we maintain a dynamic selection strategy, extracting features only for text-related key tiles. This significantly reduces memory overhead while preserving efficiency in multi-turn scenarios.

\section{Detailed Efficiency Calculation}

\textbf{Detailed calculation of vision tokens.} For vision tokens number calculation, a 4,000×4,000 image generates thumbnail view and 144 image tiles after processing with anyres-p144, resulting in $(144+1)\times576=83,520$ vision tokens. Among these, the tokens from the image tiles are downsampled using bilinear interpolation in the anyres strategy, ultimately yielding $144\times144+576=21,312$ that are fed into the LLM. This accounts for the number of vision tokens reported for anyres-p144, FastV, and PyramidDrop. For PrunMerge++ and VisionZip, we calculate the number of vision tokens based on their respective compression strategies.

For our method, we fix a 4-layer pyramid for the 4,000×4,000 input. Assuming that the thumbnail view and image tiles of the first three layers are fully utilized, this generates $(1+9+36)\times576=26,496$ vision tokens. For the fourth layer of the DIP, as our strategy dynamically selects image tiles based on the output of the RFM, we computed the average number of image tiles generated in the fourth layer for all images close to 4,000×4,000 in the datasets. This average is 50, resulting in $50\times576=28,800$ vision tokens. Therefore, the total number of vision tokens processed by the vision encoder in our method is $26,496+28,800=55,296$.  After token pruning with the ratio 0.25, the token number to LLM is $50\times144\times0.25+576=2,376$.

\textbf{Detailed calculation of TFLOPs.}
We follow PyramidDrop to calculate the TFLOPs during inference. For PyramidDrop and FastV, we compute the TFLOPs by adapting the formulas provided in their respective papers, adjusting the number of vision tokens accordingly. %

For our method, we account for the 4 layers in RFM and the 28 layers in Qwen2-7B. The FLOPs of the multi-head attention and the feedforward network modules are represented as $4nd^2+2n^2d+2ndm$, where $n$ is the number of vision tokens fed into the LLM, and $d,m$ are 3584 and 18944 in Qwen2-7B, respectively. The number of vision tokens input to RFM in DIP is calculated as $9\times144+576=1,872$ (layers 1-2), $36\times144+576=5,760$ (layer 3), and $50\times144+576=7,776$ (layer 4), respectively. Based on these values, the FLOPs $\mathcal{F}$ of our method is computed as follows:

\begin{align}
\mathcal{F} &= 4 \cdot \Big( 4 \cdot 1872 \cdot d^2 + 2 \cdot 1872^2 \cdot d + 3 \cdot 1872 \cdot d \cdot m \Big) \nonumber \\
&\quad + 4 \cdot \Big( 4 \cdot 5760 \cdot d^2 + 2 \cdot 5760^2 \cdot d + 3 \cdot 5760 \cdot d \cdot m \Big) \nonumber \\
&\quad + 4 \cdot \Big( 4 \cdot 7776 \cdot d^2 + 2 \cdot 7776^2 \cdot d + 3 \cdot 7776 \cdot d \cdot m \Big) \nonumber \\
&\quad + 28 \cdot \Big( 4 \cdot 2376 \cdot d^2 + 2 \cdot 2376^2 \cdot d + 3 \cdot 2376 \cdot d \cdot m \Big) \nonumber \\
&= 36.61 \, T
\end{align}

\section{More Experimental Results}
\label{appendix_c}

\textbf{Different distillation losses.} 
We explored different combinations of distillation losses, as shown in Tab.~\ref{tab:ablation_distill_loss}. It can be observed that the KL loss plays a crucial role, while the MSE loss applied to the high-resolution vision tokens also contributes to performance improvement.

\begin{table}[!h]
\centering
\small
\begin{tabular}{cccccc}
\toprule
\textbf{KL} & \textbf{MSE} & \textbf{Color} & \textbf{Count} & \textbf{Pos} & \textbf{Acc}  \\
\midrule
$\checkmark$ & & 44.70 & 29.04 & 49.56 & 41.20 \\
 & $\checkmark$ & 42.07 & 30.75 & 46.14 & 39.73  \\
$\checkmark$ &  $\checkmark$ &  44.70 & 31.00 & 49.72 & \textbf{41.89}   \\
\bottomrule
\end{tabular}
\caption{Ablation study on different losses used in attention distillation, under LLaVA-Next-Qwen2.}
\label{tab:ablation_distill_loss}
\vspace{-5pt}
\end{table}

\textbf{Different pruning ratios.} As shown in Tab.~\ref{tab:appendix_prune_ratios}, we employ multiple higher-resolution grids under the anyres strategy: anyres-p36 and anyres-p49. Then we conduct experiments with different pruning rates based on RFM-based pruning. The results indicate that the performance of the anyres baseline degrades as the supported image size increases, whereas our pruning method achieves consistent improvements. We attribute this to the fact that larger image sizes introduce more irrelevant background information, while our method effectively drops unnecessary vision tokens.

\begin{table}[!ht]
\centering
\small
\setlength{\tabcolsep}{4pt} 
\begin{tabular}{l | cccccc}
\toprule
\textbf{Method} & \textbf{\makecell{Drop\\ Ratio}}  & \textbf{Color} & \textbf{Count}  & \textbf{Pos} & \textbf{Acc} \\
\midrule
anyres-p36 & 0\% & 42.31 & 29.77 &  44.71 & 39.00 \\
\multirow{4}{*}{\textit{w/ prune (Ours)}} & 25\% & 42.71 & 30.10 & 45.19 & 39.41 \\
  & 50\% & 42.79 & 31.40 & 47.49  & \textbf{40.64} \\
 & 75\%  &  41.75 & 30.91 & 48.45 & 40.45 \\  \hline
anyres-p49 & 0\% & 40.00 & 30.34 & 44.95 &  38.50 \\
\multirow{4}{*}{\textit{w/ prune (Ours)}} & 25\% & 40.72 & 30.91 &  46.38 & 39.41 \\
  & 50\% & 40.64 & 31.65 & 48.61  & \textbf{40.37} \\
 & 75\%  &  41.75 & 30.51 &  48.37 & 40.29 \\ 
\bottomrule
\end{tabular}
\caption{Ablation study on different prune ratios with LLaVA-Next-Qwen2 with larger resolutions, when pruning vision tokens based on RFM results without DIP. ``pX" indicates the max number is X for the pre-defined
grids.}
\label{tab:appendix_prune_ratios}
\end{table}

\textbf{Different high-resolution vision token process.}
Additionally, for the DIP layers that have already been traversed (i.e., the layers before the final pruning layer), we attempt to prune their vision tokens as well. Specifically, we concatenate the pruned vision tokens from all traversed DIP layers, as shown in Tab.~\ref{tab:appendix_concat}, the performance slightly decreases, possibly due to interference caused by the mixed resolutions of vision tokens across multiple hierarchical DIP levels.

\begin{table}[!h]
\centering
\small
\begin{tabular}{ccccc}
\toprule
\textbf{Setting} & \textbf{Color} & \textbf{Count} & \textbf{Pos} & \textbf{Acc}  \\
\midrule
concat & 44.22 &  31.48 & 48.21 & 41.39 \\
select & 44.70 &  31.00  &  49.72  &  41.89 \\
\bottomrule
\end{tabular}
\caption{Ablation study on different high-resolution vision token processing, under LLaVA-Next-Qwen2. ``concat" means concatenating all pruned vision tokens from all traversed DIP layers as high-resolution part. ``select" means selecting the pruned tokens of the final pruning layer as high-resolution part, which is employed in the main paper.}
\label{tab:appendix_concat}
\vspace{-6pt}
\end{table}

\begin{table}[!h]
\centering
\small
\begin{tabular}{cccccc}
\toprule
\textbf{Prune} &  \textbf{Fus. Text} & \textbf{Color} & \textbf{Count} & \textbf{Pos} & \textbf{Acc} \\
\midrule
 $\checkmark$ & $\times$ & 41.91 & 26.02 & 49.01 & 39.09 \\
 $\checkmark$ & $\checkmark$  & 40.96 & 31.97 & 44.63 & 39.25 \\
 $\times$ &$\times$ & 44.70 &  31.00  &  49.72  &  41.89 \\
\bottomrule
\end{tabular}
\caption{Ablation study on training with vision token pruning and text fusion. ``Fus. Text" refers to replacing the original text tokens with the text portion (i.e., fusion text) output by the RFM when feeding into the LLM.}
\label{tab:appendix_train_prune}
\end{table}

\textbf{Training with pruning.}
We further explore the effects of directly applying pruning during the SFT stage, as shown in Tab.~\ref{tab:appendix_train_prune}. During training, the pruning ratio is set to 0.75, and the LLM observes the pruned vision tokens. Additionally, we experiment with feeding the text portion (i.e., fusion text) of the hidden states output by the RFM to the LLM, represented as ``Fus. Text" in Tab.~\ref{tab:appendix_train_prune}. The results indicate that introducing token pruning during training leads to a performance drop. We think this is because the model lacks access to complete image information, which impairs its ability to accurately perform text-aware region localization, resulting in inferior performance compared to standard SFT.

It is important to note that our method cannot utilize the fusion text setting in Tab.~\ref{tab:appendix_train_prune}. Because under such a setting, during training, the LLM receives fusion text along with full vision tokens as input, whereas during inference, the LLM receives fusion text along with pruned vision tokens. This creates an inconsistency between training and inference.

\textbf{Different layer-pairs in distillation.} 
We explore the impact of different RFM-LLM layer-pairs used for distillation, as shown in Tab.~\ref{tab:appendix_distill_layers}. Specifically, for the 28-layer Qwen2-7B, we observed that the text-related attention localization is most accurate in its deep layers (approximately layers 14–24) when answering questions about both the global content and local details of the large RSIs.

\begin{table}[tbp]
\centering
\small
\setlength{\tabcolsep}{4pt} 
\begin{tabular}{cccccc}
\toprule
\textbf{\makecell{RFM \\Layers}}  & \textbf{\makecell{LLM \\Layers}} & \textbf{Color} & \textbf{Count} & \textbf{Pos} & \textbf{Acc}  \\
\midrule
3 & \text{[1,8,14]} & 41.35 & 28.30 & 47.89 & 39.27  \\
4 & \text{[1,5,11,14]} & \textbf{44.70} &  31.00 & \textbf{49.72} & \textbf{41.89}   \\
4 & \text{[1,7,13,18]} &  40.88 & \textbf{32.63} & 49.48 & 41.06  \\
4 & \text{[1,7,14,20]} & 41.83 & 28.22 & 47.89 & 39.41 \\
5 & \text{[1,5,10,12,14]} & 43.98 & 29.36 & 48.13 & 40.58   \\
6 & \text{[1,3,6,9,12,14]} & 42.31 & 31.08 & 47.18 & 40.26  \\
6 & \text{[1,4,8,12,16,20]} & 43.75 & 30.51 & 48.13 & 40.88  \\
6 & \text{[1,5,10,15,20,24]} & 42.31 & 31.08 & 47.18 & 40.26 \\
\bottomrule
\end{tabular}
\caption{Ablation study on different RFM-LLM layer-pair configurations in MME-RealWorld-RS, with LLaVA-Next-Qwen2. ``RFM Layers" means the number of layers in RFM.}
\label{tab:appendix_distill_layers}
\end{table}

From Tab.~\ref{tab:appendix_distill_layers}, it can be observed that as the number of LLM layers for distillation increases, it becomes more challenging for the RFM to learn precise text-aware localization capabilities. Although increasing the number of layers in the RFM can enhance its learning ability, it also raises the cost of training and inference. Moreover, the RFM doesn't need to possess highly accurate localization capabilities in the shallow layers of DIP, it only needs to provide rough positions to index image tiles of the next DIP layer. Additionally, when used for pruning, the vision tokens from key image tiles already narrow down the scope, making it sufficient to recognize general background information. Furthermore, retaining a certain number of context tokens can actually be beneficial for certain types of questions.

\begin{table*}[ht!]
\centering
\begin{tabular}{l | c c c c c}
\hline
Pruning Guidance & LRS-FAIR & LRS-Bridge & LRS-STAR & DIOR-RSVG & RRSIS-D \\ 
\hline
Teacher LLM Attn.  &  53.03 &	58.66 &	56.73     &  74.81   & 73.88                \\ \hline
CLIP Sim.        &   23.87   & 16.73   & 19.82       &     29.10  &   27.53             \\
RemoteCLIP Sim.    &    37.04 &	22.99  &  32.21   &  43.12  &    41.81         \\
RFM Attn. (Ours)      &   \textbf{47.89} & \textbf{49.08} & \textbf{47.01}  &   \textbf{64.76}   &  \textbf{61.17}   \\  
\hline
\end{tabular}
\caption{\small{Localisation recall (\%) of different pruning guidance.}}
\label{tab:location_acc_baselines}
\end{table*}

\begin{table*}[ht!]
\centering
\linespread{0.7}
\begin{tabular}{@{}l|cc|cccc@{}}
\hline
Setting & MME-RW-RS & FPS & LRS-FAIR & LRS-Bridge & LRS-STAR & FPS \\
\hline
LLaVA-Next-p25 & 39.65 & 0.188 & 20.99 & 36.38 & 26.18 & 0.176 \\
\textit{w/ CLIP Sim.} & 36.86 & 0.171 & 18.12 & 32.30 & 24.46 & 0.162 \\
\textit{w/ RemoteCLIP Sim.} & 38.77 & 0.148 & 20.36 & 34.24 & 25.32 & 0.139\\
\textit{w/ RFM (Ours)} & \textbf{41.28} & 0.165 & \textbf{21.65}  & \textbf{37.55} & \textbf{26.83}& 0.157\\
\hline
\end{tabular}
\linespread{0.92}
\caption{\small{VQA accuracy (\%) and FPS with different token pruning guidance methods. FPS on LRS-VQA averaged across 3 datasets.}}

\label{tab:token_pruning_baselines} 
\end{table*}

\textbf{Detailed comparison with baselines.} We provide a more comprehensive comparison with two simple but vital baselines: CLIP-L14 and RemoteCLIP-L14. Specifically, under identical anyres or DIP settings, we partition the image using a sliding-window approach (336×336 for CLIP and 224×224 for RemoteCLIP). Then we compute the similarity map between the input text and the image features for each image tile, which are subsequently stitched together to form a complete heatmap that guides the token pruning.

Tab.~\ref{tab:location_acc_baselines} presents a comparison of localization accuracy against these baselines across three datasets. While conceptually simpler, these baselines struggle due to their limited capacity to understand complex referring expressions and the lack of global perception inherent in the sliding-window mechanism. This highlights the importance of distilling knowledge from the LLM's attention, which enables our RFM to grasp complex semantics.

Furthermore, Tab.~\ref{tab:token_pruning_baselines} shows the performance and FPS comparison under the same pruning setting (LLaVA-Next-p25). The accuracy trends observed here are largely consistent with the localization accuracy results. It is worth noting that although RemoteCLIP enhances performance on remote sensing imagery, its smaller input size of 224×224 necessitates partitioning the image into more tiles, which adversely affects inference speed on large images.

\begin{table*}[!h]
\centering
\begin{tabular}{l |c|ccccccc}
\toprule
\textbf{Method}                 & \textbf{Max Res.}    & count & category & shape & status & reasoning & rural/urban & \textbf{OverAll} \\ \hline
LLaVA1.5   & 336×336       & 10.50 & 13.44    & 7.37  & 7.75   & 25.25     & 48.25       & 18.76   \\
SLiME    & 672×1,008      & 14.25 & 9.82     & 8.07  & 9.25   & 24.50     & 36.75       & 17.11   \\
SEAL                            &    -      & 7.00  & 12.50    & 3.50  & 20.50  & 28.75     & 55.50       & 21.29   \\
LLaVA-FlexAttn &  1,008×1,008        & 9.50  & 10.59    & 6.32  & 20.75  & 24.00     & 46.25       & 19.57   \\
MGM-HD  &     1,536×1,536   & 15.50 & 12.66    & 9.47  & 6.00   & 23.75     & 40.00       & 17.90   \\
LLaVA-UHD-v2      &     672×1,008   & 17.50 & 11.63    & 9.04  & 23.00  & 26.25     & 49.50       & 22.82   \\

IXC-2.5 &  4,096×4,096 & 22.75 & 15.25 & 15.50 & 22.00 & 26.50 & 49.50 & 25.25 \\
LLaVA-OV     &    2,304×2,304   & 16.25 & 19.90    & 8.77  & 12.75  & 27.25     & 38.75       & 20.61   \\
Qwen2-VL        &  3,333×3,333 & 22.50 & 15.25    & 12.28 & 10.00  & 24.25     & 58.50       & 23.80   \\ \hline
Geochat    &    504×504      & 13.50 & 8.01     & 14.04 & 3.50   & 19.75     & 62.25       & 20.18   \\
RSUniVLM              &  336×336    & 21.00 & 11.37    & 15.98 & 2.00   & 25.00     & 50.75       & 21.02   \\ \hline
Claude-3.5-Sonnet               &    -    & 11.75 & 4.12     & 16.84 & 1.50   & 15.00     & 28.50       & 12.95   \\
Gpt-4o-mini                     &    -      & 12.75 & 11.37    & 11.37 & 12.50  & 19.75     & 44.25       & 18.67   \\
Gpt-4o                          &    -     & 16.00 & 13.44    & 14.98 & 18.00  & 24.00     & 46.50       & 22.15  \\
\bottomrule
\end{tabular}
\caption{Detailed results on LRS-FAIR.}
\label{tab:detail_fair_result}
\end{table*}

\begin{table*}[!h]
\centering
\begin{tabular}{l |c|ccccc}
\toprule
\textbf{Method}              & \textbf{Max Res.}      & count & background & color & rural/urban & \textbf{OverAll} \\ \hline
LLaVA1.5            & 336×336     & 6.50  & 18.37      & 38.79 & 59.13       & 30.70   \\
SLiME              & 672×1,008    & 13.50 & 18.78      & 34.55 & 61.51       & 32.09   \\
SEAL                &  -  & 0.00  & 31.50      & 36.00 & 71.50       & 34.75   \\
LLaVA-FlexAttn     &  1,008×1,008        & 4.50  & 17.96      & 37.58 & 59.92       & 29.99   \\
MGM-HD              &    1,536×1,536    & 12.25 & 18.78      & 52.73 & 59.92       & 35.92   \\
LLaVA-UHD-v2        & 672×1,008   & 5.00  & 17.55      & 44.24 & 63.49       & 32.57   \\
IXC-2.5 &  4,096×4,096 & 14.50 & 20.30 & 55.34 & 63.49 & 38.41 \\
LLaVA-OV            &   2,304×2,304  & 4.50  & 20.82      & 57.58 & 57.54       & 35.11   \\
Qwen2-VL            &  3,333×3,333     & 15.50 & 20.41      & 57.58 & 59.00       & 38.12   \\ \hline
Geochat             &     504×504        & 8.75  & 11.84      & 22.42 & 55.16       & 24.54   \\
RSUniVLM            &   336×336     & 10.25 & 13.06      & 43.64 & 63.49       & 32.61   \\ \hline
Claude-3.5-Sonnet&-& 5.25  & 11.43      & 33.33 & 56.75       & 26.69   \\
Gpt-4o-mini         &    -    & 4.75  & 17.96      & 50.97 & 54.29       & 31.99   \\
Gpt-4o              &   -   & 14.75 & 18.37      & 43.03 & 51.19       & 31.84    \\
\bottomrule
\end{tabular}
\caption{Detailed results on LRS-Bridge.}
\label{tab:detail_bridge_result}
\end{table*}

\begin{table*}[!h]
\centering
\begin{tabular}{l |c|cccccccc}
\toprule
\textbf{Method}            & \textbf{Max Res.} & count & category & color & shape & status & reasoning & rural/urban & \textbf{OverAll} \\ \hline
LLaVA1.5          &  336×336   & 10.75 & 18.67    & 36.50 & 10.00 & 11.33  & 15.67     & 55.50       & 22.63   \\
SLiME             & 672×1,008   & 12.75 & 18.33    & 41.50 & 10.50 & 12.67  & 16.00     & 49.17       & 22.99   \\
SEAL              &  -   & 0.25  & 20.50    & 33.50 & 10.00 & 10.50  & 17.75     & 56.50       & 21.29   \\
LLaVA-FlexAttn    &  1,008×1,008  & 11.50 & 17.00    & 36.17 & 8.67  & 13.33  & 14.83     & 57.83       & 22.76   \\
MGM-HD            &  1,536×1,536    & 15.50 & 12.66    & 49.50 & 18.75 & 10.30  & 10.18     & 24.00       & 20.13   \\
LLaVA-UHD-v2      &  672×1,008   & 16.25 & 18.67    & 43.50 & 15.67 & 14.83  & 16.17     & 57.50       & 26.08   \\
IXC-2.5 &  4,096×4,096 & 15.75 & 23.50 & 48.00 & 16.50 & 13.33 & 17.50 & 56.50 &27.30  \\
LLaVA-OV          &  2,304×2,304   & 9.75  & 25.33    & 51.00 & 14.00 & 10.17  & 18.83     & 53.50       & 26.08   \\
Qwen2-VL          &   3,333×3,333     & 19.25 & 22.83    & 46.50 & 11.17 & 13.00  & 18.33     & 64.00       & 27.87   \\ \hline
Geochat           &  504×504     & 13.50 & 8.01     & 24.75 & 10.75 & 5.42   & 14.04     & 19.75       & 13.75   \\
RSUniVLM          &  336×336      & 8.00  & 13.50    & 51.67 & 25.17 & 4.50   & 14.00     & 56.17       & 24.72   \\ \hline
Claude-3.5-Sonnet &  -    & 6.34  & 6.00     & 41.67 & 1.67  & 12.33  & 3.00      & 22.00       & 13.29   \\
Gpt-4o-mini       &  -    & 10.78 & 20.67    & 40.67 & 15.17 & 14.50  & 20.33     & 58.83       & 25.85   \\
Gpt-4o            &   - & 11.78 & 21.50    & 48.17 & 23.83 & 12.50  & 20.50     & 53.50       & 27.40    \\
\bottomrule
\end{tabular}
\caption{Detailed results on LRS-STAR.}
\label{tab:detail_star_result}
\end{table*}

\textbf{Detailed results on LRS-VQA.} The complete leaderboards on the three parts of LRS-VQA are shown in Tab.~\ref{tab:detail_fair_result}, Tab.~\ref{tab:detail_bridge_result} and Tab.~\ref{tab:detail_star_result}, respectively. Notably, on questions that require more global-scale perception capabilities, such as rural/urban classification, high-resolution LVLMs do not necessarily outperform low-resolution LVLMs. Additionally, the language preference inherent in the LVLM itself can influence its performance when answering open-ended questions, as it must precisely describe the corresponding vocabulary or its synonyms. Overall, LVLMs like Qwen2-VL, LLaVA-OV, and IXC-2.5, which are trained on large datasets and utilize higher resolutions, also demonstrate strong performance in large RSI perception tasks.

\end{appendices}

\end{document}